\documentclass{article}

\usepackage[dblblindworkshop, final]{neurips_2025}

\workshoptitle{Continual and Compatible Foundation Model Updates (CCFM)}

\usepackage[utf8]{inputenc} % allow utf-8 input
\usepackage[T1]{fontenc}    % use 8-bit T1 fonts
\usepackage{hyperref}       % hyperlinks
\usepackage{url}            % simple URL typesetting
\usepackage{booktabs}       % professional-quality tables
\usepackage{amsfonts}       % blackboard math symbols
\usepackage{nicefrac}       % compact symbols for 1/2, etc.
\usepackage{microtype}      % microtypography
\usepackage{xcolor}         % colors
\usepackage{graphicx} % Required for inserting images
\usepackage{amsmath} 
\usepackage{multirow}
\usepackage{tabularx}
\usepackage{color}
\usepackage{soul}
\usepackage{amssymb}
\usepackage{mathtools}
\usepackage{amsthm}
\usepackage{enumitem}
\usepackage{array}
\usepackage{wrapfig}
\usepackage{svg}
\usepackage{pgfplots}
\usepackage{layouts}
\usepackage{subfig}
\usepackage{placeins}
\usepackage{bm}
\usepackage{algorithm}
\usepackage{algpseudocode}
\usepackage{printlen}
\usepackage{tikz}
\usepackage{mdframed}
\usetikzlibrary{fit,calc}
\usepackage{chngcntr}
\usepackage{selectp}
\usepackage{chngcntr} % For resetting counters
\usepackage{apptools} % For appendix tools
\usepackage{tcolorbox}

\usepackage{amsthm}

\usepackage{colortbl}  % Optional, for additional color table functionality

\usepackage{todonotes}

\title{Balancing Synthetic Data and Replay for\\Enhancing Task-Specific Capabilities}

\author{
  Urs Spiegelhalter$^{1}$ \quad
  Jörg K.H. Franke$^{1,2,3, 4}$ \quad
  Frank Hutter$^{1,2,5}$ \quad
  \\ \\
  $^1$University of Freiburg \quad
  $^2$ELLIS Institute Tübingen \quad \\
  $^3$Open-Sci Collective \quad
  $^4$LAION \quad
  $^5$Prior Labs \quad
}

\begin{document}

\maketitle

\begin{abstract}
Adapting language models to new tasks through continued pretraining faces a fundamental trade-off: models must learn new capabilities while avoiding catastrophic forgetting of existing knowledge. While prior work has studied synthetic data generation techniques, the optimal replay ratios for balancing task performance and knowledge retention under computational constraints remain poorly understood. We present a comprehensive empirical study investigating the interplay between replay ratio configuration and computational budget when adapting language models to new tasks. Using the bAbI reasoning tasks as our target objective, we apply synthetic data generation and systematically evaluate different total token budgets and replay ratio configurations. We analyze their effects on both task mastery and general knowledge retention. Our experiments reveal an optimal configuration that balances task-specific performance with general knowledge retention. Based on our findings, we provide empirically-grounded guidelines for selecting replay ratios based on computational budget, enabling practitioners to achieve strong task adaptation with significantly reduced training costs.
\end{abstract}

\section{Introduction}

Foundation models~\citep{grattafiori2024llama3, gemmateam2025gemma3, deepseekai2025deepseekv3} face a fundamental challenge in continual learning: how to acquire new task-specific capabilities without degrading their general performance~\citep{shi2024continual, luo2025empirical}. This challenge is especially prominent when adapting models to specialized tasks where training data is limited but task mastery is critical. While synthetic data generation and experience replay have emerged as promising approaches~\citep{yang2024synthetic, jiang2025synthesize, ibrahim2024simple}, the optimal balance between these strategies remains underinvestigated, especially under computational constraints.

In this work, we systematically analyze the impact of replay on different token budgets. Our approach combines two key components: (i) synthetic data generation that expands limited task-specific datasets through controlled diversity expansion, and (ii) systematic analysis of replay ratios across different computational budgets to identify configurations that balance task mastery with general knowledge retention.

For synthetic data generation, we leverage a task suite that offers structured reasoning patterns suitable for controlled augmentation. The bAbI reasoning tasks~\citep{weston2015towards} provide an ideal testbed for studying this balance, as they require precise reasoning capabilities that can be systematically evaluated. 

Our key contributions are:
\begin{itemize}[leftmargin=2em, itemsep=0pt]
\item A systematic evaluation of the interplay between synthetic data scale and replay percentage in total token constrained continual learning.
\item Evidence that more than $5\text{-}10\%$ replay is not necessary for general knowledge retention. This configuration provides the optimal trade-off between task mastery and general knowledge retention.
\item Evidence that synthetic data diversity is crucial for task mastery: training with multiple epochs on smaller, less diverse datasets performs significantly worse than training on unique diverse samples with the same total token budget.
\end{itemize}

\section{Related Work}

\paragraph{Continual Learning in Foundation Models}
Foundational work has highlighted the challenge of updating foundation models without forgetting~\citep{ramasesh2022effect, scialom2022fine}. Experience replay, where previous training data is mixed with new task data, remains one of the most effective approaches~\citep{chaudhry2019tiny}. However, determining optimal replay ratios typically requires expensive empirical search. Recent work by \citet{ibrahim2024simple} investigates replay ratio effects in the context of broad domain adaptation, such as adapting models to new languages. In contrast, our work focuses on narrow task-specific adaptation using synthetic data.

\paragraph{Continued Pretraining with Synthetic Data}
Recent work~\citep{yang2024synthetic, jiang2025synthesize, chen2024diversity} uses synthetically enriched data from a small domain to perform continued pretraining. While \citet{chen2024diversity} explicitly explores synthetic dataset diversity and \citet{jiang2025synthesize} explores synthetic dataset sizes, they do not investigate catastrophic forgetting or general knowledge retention in their experiments.

\section{Method}

\subsection{Synthetic Data Generation}

We use the bAbI tasks~\citep{weston2015towards} as our basis for task-specific continued pretraining since they offer a wide variety of tasks. However, while the legacy code base\footnote{https://github.com/facebookarchive/bAbI-tasks} of the original tasks exists, limited diversity in the entities used poses challenges for robust generalization. We address this limitation through controlled synthetic expansion by introducing novel entities. This enables us to generate a few million unique samples per task, instead of only a few thousand.
We provide more details on the bAbI tasks and our synthetic expansion in Appendix~\ref{app:babi}.

Additionally, we investigate the impact of synthetic data diversity. Therefore, we use two bAbI datasets variants: \textbf{bAbI-Original}: the standard bAbI dataset containing 1,000 unique samples per task, generated from the original code base with a limited entity set, and \textbf{bAbI-Synthetic}: our synthetically expanded version containing up to hundreds of thousands of unique samples per task. For training, we adopt different sampling strategies based on the dataset variant. With bAbI-Original, we perform multiple epochs over the 1,000 samples per task until reaching the target token budget. For bAbI-Synthetic, we sample unique samples without replacement up to the target token budget, leveraging the dataset's larger diversity to avoid repetition.

\subsection{Optimal Configuration}
A configuration consists of a total token budget and a replay percentage, where the replay percentage specifies the proportion of tokens allocated to task-specific data. For acquiring new task expertise while preserving general capabilities, we prioritize: (i) minimizing computational cost via total token budget, and (ii) maximizing task-specific data allocation through minimal replay percentage, enabling learning of more complex tasks within the same budget.
We define an optimal configuration as one that minimizes both replay percentage and total token budget while maintaining the combined benchmark score: the sum of task-specific performance and average change in general benchmarks. At this optimum, increasing either parameter yields only marginal improvements in the combined score, indicating efficient resource utilization.

\section{Experiments}

\subsection{Experimental Setup}
We conduct experiments using the bAbI tasks as our task to learn and the DCLM-Edu~\citep{dclm-edu} pretraining dataset as the replay dataset. Our base model, the SmolLM2-1.7B LLM~\citep{allal2025smollm2}, achieves a performance of $54.9\%$ on the bAbI tasks, leaving room for improvement.

Therefore, we conduct experiments across a comprehensive grid of configurations, varying both the total token budget and replay percentage. Specifically, we evaluate five token budgets \{1e7, 1e7.5, 1e8, 1e8.5, 1e9 tokens\} and five replay percentages \{$5\%$, $10\%$, $15\%$, $20\%$, $25\%$\}, resulting in 25 distinct configurations. We evaluate these configurations on both bAbI-Original and bAbI-Synthetic.

Hyperparameters were determined through a preliminary grid search using 50\% replay and 100k bAbI-Synthetic samples per task. Our final configurations use a peak learning rate of 0.0005, a batch size of 256, with a linear warm-up for the first $10\%$ of training followed by cosine decay. We provide more details on the training setup, hyperparameter search, and dataset statistics in Appendix~\ref{app:train_hp}.

\subsection{Evaluation Protocol}

We evaluate models on two complementary metrics. First, we evaluate bAbI task performance by measuring accuracy on a held-out bAbI test set containing completely unseen entities. We evaluate using 5-shot prompting with examples from the training split to ensure consistent answer formatting. Second, we assess general performance through average performance across HellaSwag~\citep{zellers2019hellaswag}, ARC-Easy~\citep{clark2018arc}, PIQA~\citep{bisk2020piqa}, MMLU~\citep{hendrycks2021mmlu}, CommonsenseQA~\citep{talmor2019commonsenseqa}, WinoGrande~\citep{sakaguchi2020winogrande}, OpenBookQA~\citep{mihaylov2018openbookqa}, and MathQA~\citep{amini2019mathqa} benchmarks, evaluated with the Language Model Evaluation Harness~\citep{eval-harness}. The combined score is calculated by the sum of the bAbI test set accuracy and the average change of the general performance benchmarks.

We initially also included GSM8K~\citep{cobbe2021gsm8k} in the general performance benchmarks but removed it from our main analysis due to DCLM-Edu's inefficient performance retention on GSM8K. However, we observe that augmenting the replay mixture with a small proportion of mathematical data effectively addresses this degradation. We provide more details in Appendix~\ref{app:gsm8k}.

\subsection{Main Results}

\begin{figure*}[t]%
    \centering
    \includegraphics[width=1.0\textwidth]{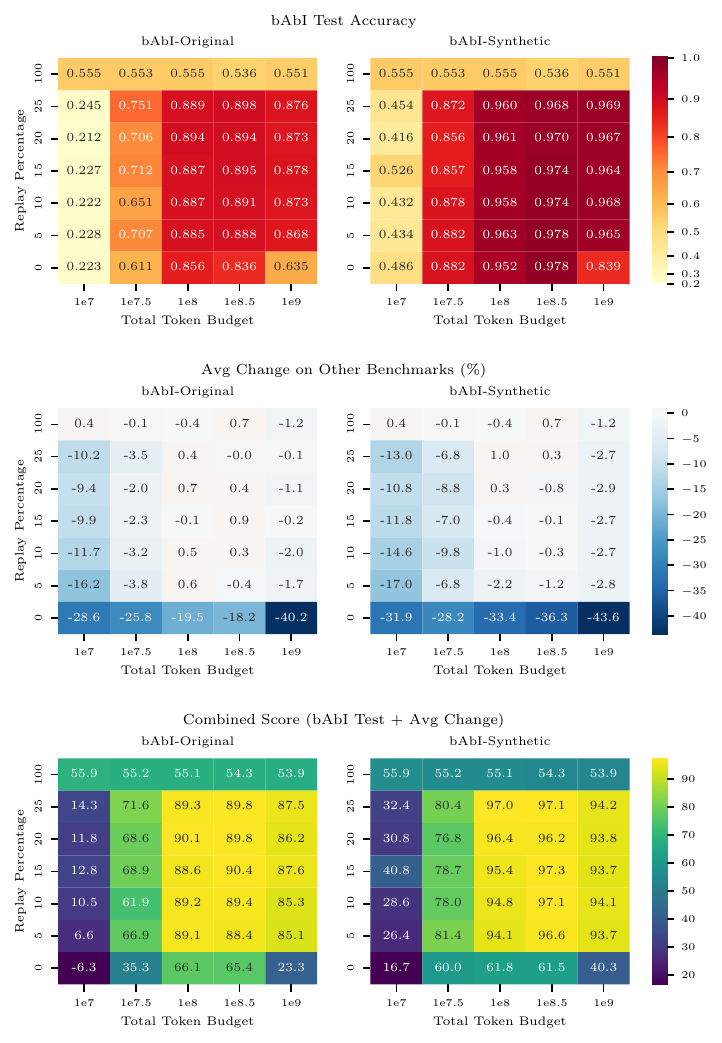}
    \caption{Full analysis of the interplay between different total token budgets and replay percentages. The left column shows results for bAbI-Original and the right column shows results for bAbI-Synthetic. 100$\%$ replay is exclusively training on DCLM-Edu up to that total token budget. The combined scores indicate the most valuable configurations for task mastery and general knowledge retention.}
    \label{fig:main_results_heat_map}
\end{figure*}

Figure~\ref{fig:main_results_heat_map} shows the results of the full analysis of the interplay between different total token budgets and replay percentages on bAbI-Original and bAbI-Synthetic. Our results reveal several key insights:

\paragraph{Total Token Budget}
Training beyond 1e8.5 total tokens provides no further improvements in downstream bAbI task performance. Performance plateaus at this point for both datasets. We can even observe a small decline in bAbI task performance and general knowledge retention with 1e9 total tokens.

\paragraph{General Knowledge Retention}
Between $5\%$ and $10\%$ replay is sufficient for maintaining general knowledge. Increasing replay above $10\%$ provides no meaningful additional benefit while reducing task-specific training data or increasing computational cost. Detailed performance breakdowns for the individual benchmarks are provided in Appendix~\ref{app:benchmarks}.

\paragraph{Task Diversity Impact}
bAbI-Synthetic achieves superior task performance and exhibits stable general knowledge retention patterns across different replay configurations. The substantial performance gap between bAbI-Synthetic and bAbI-Original underscores that data diversity is crucial for task mastery. Training with multiple epochs on smaller datasets performs significantly worse than training on unique diverse samples with the same token budget, consistent with \citet{chen2024diversity}.

\FloatBarrier

\section{Practical Guidelines}
For practitioners adapting models to new tasks, we recommend: (i) Initial hyperparameter search using $20\%$ replay with total token budget of 1e8. (ii) Grid search over token budgets \{1e8, 1e8.5, 1e9\} and replay ratios \{$5\%$, $10\%$, $15\%$\} to identify the configuration that best balances task performance and general knowledge retention for the specific requirements. Based on our findings, configurations outside of these ranges perform worse or provide minimal benefit while increasing computational cost.

\section{Discussion}
The emergence of a performance ceiling at 1e8.5 tokens, coupled with slight degradation at 1e9 tokens, suggests that continual learning scenarios require careful consideration of training duration. The sufficiency of $5\text{-}10\%$ replay for maintaining general knowledge demonstrates that catastrophic forgetting can be prevented with minimal computational overhead, enabling efficient adaptation without extensive rehearsal of prior training data. Our results also reveal that data diversity fundamentally changes training dynamics. The bAbI-Synthetic experiments show that unique samples consistently outperform repeated epochs, with important implications for practitioners with limited original data: investing in synthetic diversity generation yields better performance than training longer on existing data.

While our study focuses on a single task type (reasoning) and model size (1.7B parameters), and our synthetic generation method relies on rule-based templates, which may not generalize to more complex tasks, these findings provide actionable guidelines for continual learning scenarios. Different tasks or model sizes may exhibit different optimal configurations, and extending this analysis to other task types, model scales, and synthetic generation methods would strengthen the generalizability of our guidelines.

\section{Conclusion}
This work establishes empirically validated guidelines for efficient task-specific model adaptation through systematic exploration of replay strategies and training duration. We demonstrate that $5\text{-}10\%$ replay suffices to prevent catastrophic forgetting, while training beyond 1e8.5 tokens yields no benefits and may even harm performance. Combined with our finding that diverse synthetic samples consistently outperform repeated epochs, these results enable practitioners to achieve strong task performance with minimal computational overhead. Our insights provide a foundation for making continual learning practical in resource-constrained environments where efficiency is critical.

\newpage

\section*{Acknowledgements}

This research was funded by the Deutsche Forschungsgemeinschaft (DFG, German Research Foundation) under grant number 417962828.
We acknowledge funding by the European Union (via ERC Consolidator Grant DeepLearning 2.0, grant no.~101045765). Views and opinions expressed are, however, those of the author(s) only and do not necessarily reflect those of the European Union or the European Research Council. Neither the European Union nor the granting authority can be held responsible for them. \begin{center}\includegraphics[width=0.3\textwidth]{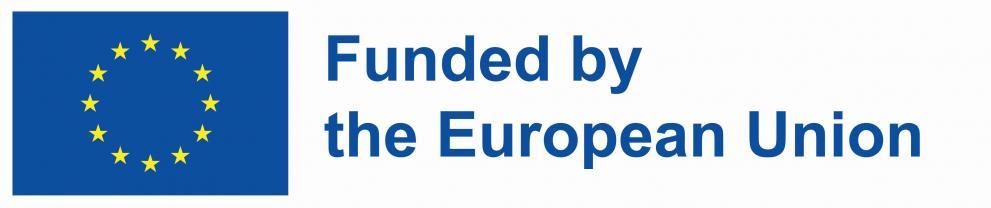}\end{center}

We gratefully acknowledge the Gauss Centre for Supercomputing e.V. for funding this work by providing computing time through the John von Neumann Institute for Computing (NIC) on the supercomputer JUWELS Booster at Jülich Supercomputing Centre (JSC).

\FloatBarrier

\bibliography{bib/refs}
\bibliographystyle{plainnat}

%%%%%%%%%%%%%%%%%%%%%%%%%%%%%%%%%%%%%%%%%%%%%%%%%%%%%%%%%%%%
\newpage
\appendix

\section*{\centering Appendix}

\section{bAbI Tasks and Synthetic Expansion}
\label{app:babi}

\subsection{Overview of bAbI Tasks}
The bAbI tasks~\citep{weston2015towards} consist of 20 question-answering tasks designed to test different aspects of text understanding and reasoning. Each task focuses on a specific reasoning capability. The tasks are presented in a simple, controlled language format to isolate specific reasoning challenges while minimizing linguistic complexity.

The dataset was originally designed with a limited vocabulary and entity set, making it suitable for systematic evaluation but potentially limiting in terms of generalization. Each task consists of a context (a series of statements) followed by questions about that context. The model must reason over the provided information to answer correctly.

We present examples of all 20 bAbI tasks below. Task complexity and difficulty scale with the number of statements preceding each question. The examples shown are picked from the lower complexity levels with very few statements.

\definecolor{lightgray}{gray}{0.95}

\newcommand{\taskexamplefontsize}{\small}

\newcommand{\taskexample}[2][]{%
 \begin{center}
 \begin{tcolorbox}[
  colback=lightgray,
  colframe=gray!75,
  boxrule=0.5pt,
  arc=2mm,
  left=3mm,
  right=3mm,
  top=2mm,
  bottom=2mm,
  width=0.75\textwidth,
  title={#1}
 ]
 \taskexamplefontsize\ttfamily #2
 \end{tcolorbox}
 \end{center}
 \vspace{1mm}
}

\subsubsection{Task 1: Location Tracking - Single Supporting Fact}
\taskexample{
1 Daniel went to the office. \\
2 Mary moved to the garden. \\
3 Where is Mary?	garden	2
}

\subsubsection{Task 2: Object Location - Two Supporting Facts}
\taskexample{
1 John grabbed the apple. \\
2 Sandra grabbed the football. \\
3 John moved to the bathroom. \\
4 Where is the apple?	bathroom	1 3
}

\subsubsection{Task 3: Object Location - Three Supporting Facts}
\taskexample{
1 Sandra took the football. \\
2 John took the milk. \\
3 Daniel moved to the kitchen. \\
4 Daniel got the apple. \\
5 Daniel moved to the office. \\
6 Where was the apple before the office?	kitchen	3 4 5
}

\subsubsection{Task 4: Spatial Relationships}
\taskexample{
1 The office is west the kitchen. \\
2 The bathroom is west the office. \\
3 What is east the office?	kitchen	1
}

\subsubsection{Task 5: Object Location - Transfer Relationships}
\taskexample{
1 Sandra took the football. \\
2 John took the milk. \\
3 Daniel journeyed to the kitchen. \\
4 Daniel went to the bedroom. \\
5 Sandra moved to the bedroom. \\
6 Sandra gave Mary the football. \\
7 Who received the football last?	Mary	6
}

\subsubsection{Task 6: Location Tracking - Yes/No Questions}
\taskexample{
1 Daniel journeyed to the garden. \\
2 Daniel journeyed to the bathroom. \\
3 Is Daniel in the bedroom?	no	2
}

\subsubsection{Task 7: Counting}
\taskexample{
1 Sandra travelled to the hallway. \\
2 Daniel moved to the kitchen. \\
3 John took the milk. \\
4 How many objects is John holding?	one	3
}

\subsubsection{Task 8: Object Location - Lists}
\taskexample{
1 Daniel moved to the bathroom. \\
2 John took the apple. \\
3 John grabbed the milk. \\
4 John went to the bathroom. \\
5 What is John holding?	apple,milk	2 3
}

\subsubsection{Task 9: Location Tracking - Negation}
\taskexample{
1 John went to the garden. \\
2 Sandra moved to the office. \\
3 Daniel is no longer in the bedroom. \\
4 Sandra is in the office. \\
5 Is Daniel in the bedroom?	no	3
}

\subsubsection{Task 10: Location Tracking - Indefinite Knowledge}
\taskexample{
1 Daniel is in the hallway. \\
2 John went to the hallway. \\
3 Sandra is either in the office or in the bedroom. \\
4 John travelled to the garden. \\
5 Is Sandra in the bedroom?	maybe	4
}

\subsubsection{Task 11: Location Tracking - Basic Coreference}
\taskexample{
1 John journeyed to the bedroom. \\
2 Following that he went to the garden. \\
3 Where is John?	garden	1 2
}

\subsubsection{Task 12: Location Tracking - Conjunctions}
\taskexample{
1 John and Mary went to the garden. \\
2 Sandra and Mary went to the bedroom. \\
3 Where is John?	garden	1
}

\subsubsection{Task 13: Location Tracking - Compound Coreference}
\taskexample{
1 John and Sandra journeyed to the kitchen. \\
2 After that they journeyed to the bathroom. \\
3 Where is Sandra?	bathroom	1 2
}

\subsubsection{Task 14: Location Tracking - Time Reasoning}
\taskexample{
1 Daniel moved to the kitchen this morning. \\
2 Mary journeyed to the bathroom this evening. \\
3 John travelled to the kitchen this afternoon. \\
4 Yesterday Daniel moved to the hallway. \\
5 Where was Daniel before the kitchen?	hallway	1 4
}

\subsubsection{Task 15: Animal Fear Chains - Basic Deduction}
\taskexample{
1 Wolves are afraid of sheep. \\
2 Emily is a mouse. \\
3 Gertrude is a sheep. \\
4 Jessica is a wolf. \\
5 Winona is a cat. \\
6 Sheep are afraid of cats. \\
7 Mice are afraid of sheep. \\
8 Cats are afraid of mice. \\
9 What is Gertrude afraid of?	cat	3 6
}

\subsubsection{Task 16: Color-Animal Associations - Basic Induction}
\taskexample{
1 Brian is white. \\
2 Bernhard is yellow. \\
3 Greg is gray. \\
4 Greg is a swan. \\
5 Julius is a swan. \\
6 Brian is a rhino. \\
7 Lily is a rhino. \\
8 Julius is gray. \\
9 Bernhard is a frog. \\
10 What color is Lily?	white	1 6 7
}

\subsubsection{Task 17: Positional Reasoning}
\taskexample{
1 The blue triangle is below the red sphere. \\
2 The red sphere is to the right of the rectangle. \\
3 Is the red sphere above the blue triangle?	yes	1 2
}

\subsubsection{Task 18: Size Reasoning}
\taskexample{
1 The container is bigger than the chocolate. \\
2 The suitcase is bigger than the chocolate. \\
3 The container is bigger than the box. \\
4 The box is bigger than the suitcase. \\
5 Does the container fit in the suitcase?	no	3 4
}

\subsubsection{Task 19: Path Finding}
\label{app:babi:task19}
\taskexample{
1 The bedroom is east the hallway. \\
2 The bathroom is south the kitchen. \\
3 The office is east the bedroom. \\
4 The kitchen is east the office. \\
5 What is the path from office to bathroom?	e,s	2 4
}

\subsubsection{Task 20: Agent’s Motivations}
\taskexample{
1 Jason is tired. \\
2 Jason travelled to the bedroom. \\
3 Why did Jason go to the bedroom?	tired	1
}

The examples presented above represent the unprocessed outputs from the legacy bAbI tasks codebase\footnote{https://github.com/facebookarchive/bAbI-tasks}, with the original entities. We apply several preprocessing modifications to these raw samples to enhance their suitability as high-quality training data. For instance, in Task 19, we insert the preposition "of" following each directional indicator.

We excluded Task 20 from our training set based on preliminary experimental findings. Our initial studies revealed limited generalization capabilities, particularly when utilizing our expanded bAbI datasets. We attribute this limitation to Task 20's underlying assumption of a one-to-one correspondence between motivations and room destinations. The example provided illustrates this mapping: "tired" corresponds to "bedroom." Similarly, the dataset contains mappings such as "bored" to "garden." The model's ability to memorize these specific motivation-room pairings observed during training does not transfer to novel, unseen mappings at test time, thereby constraining its generalization performance.

To construct the bAbI training samples, we concatenated the context statements, question, and answer components. We formatted the answer using the template "The answer is <answer word(s)>". For example, a bAbI training sample looks like this: "Mary journeyed to the hallway. Mary went to the garden. Where is Mary? The answer is garden."

\subsection{bAbI Synthetic Expansion}
The entities utilized in the bAbI tasks are stored either as text files or embedded directly within the task source code. For entities defined in source code files, we extend the typical set of 4-6 entities to 100 entities to enhance diversity.

The text file storing entities for tasks 1, 2, 3, 5, 6, 7, 9, 10, 11, 12, and 13 is structured as follows:

\renewcommand{\taskexamplefontsize}{\scriptsize}
\taskexample[world\_basic.txt]{
\# locations \\
create bedroom \\
set bedroom is\_thing \\
set bedroom is\_location \\

create kitchen \\
set kitchen is\_thing \\
set kitchen is\_location \\

create garden \\
set garden is\_thing \\
set garden is\_location \\

create hallway \\
set hallway is\_thing \\
set hallway is\_location \\

create bathroom \\
set bathroom is\_thing \\
set bathroom is\_location \\

create office \\
set office is\_thing \\
set office is\_location \\

\# small (moveable) objects \\
create apple \\
set apple is\_thing \\
set apple is\_gettable \\
set apple is\_in kitchen \\

create football \\
set football is\_thing \\
set football is\_gettable \\
set football is\_in garden \\

create milk \\
set milk is\_thing \\
set milk is\_gettable \\
set milk is\_in kitchen \\

\# objects \\
create table \\
set table is\_thing \\
set table is\_in kitchen \\

\# actors \\
create John \\
set John is\_actor \\
set John is\_in kitchen \\
set John is\_god \\
set John is\_male \\

create Mary \\
set Mary is\_actor \\
set Mary is\_in bedroom \\
set Mary is\_god \\
set Mary is\_female \\

create Sandra \\
set Sandra is\_actor \\
set Sandra is\_in garden \\
set Sandra is\_god \\
set Sandra is\_female \\

create Daniel \\
set Daniel is\_actor \\
set Daniel is\_in hallway \\
set Daniel is\_god \\
set Daniel is\_male \\
}

A straightforward approach would involve directly extending this entity list. However, this method proves ineffective as numerous tasks depend on entity sampling until meeting certain conditions. As the number of entities increases, the sampling process becomes progressively more computationally expensive while simultaneously reducing diversity. To address these limitations, we implement the following methodology:

\begin{enumerate}
    \item We construct 10 distinct basic world entity files, each containing a marginally increased number of entities to maintain acceptable runtime performance. These 10 basic world entities are mutually disjoint.
    \item During task sample generation, we first randomly select one of the 10 basic worlds, then proceed with the standard code execution.
\end{enumerate}

This approach enables efficient expansion of diversity in the bAbI tasks while maintaining computational efficiency.

\FloatBarrier
\section{Training Details and Hyperparameter Search}
\label{app:train_hp}

\subsection{Training Details}
We conduct our experiments using SmolLM2-1.7B \cite{allal2025smollm2}, a 1.7B parameter model from the SmolLM2 family. This model offers several practical advantages for our experimental setup: (i) its 2048-token context length allows us to train at the model's native sequence length without truncation on our available hardware, and (ii) as a monolingual English model, the DCLM-Edu~\citep{dclm-edu} pretraining dataset provides a well-characterized distribution for controlled replay experiments.

To construct our training datasets, we combine replay data from DCLM-Edu with samples from the bAbI tasks \cite{weston2015towards}. Specifically, for a replay percentage $r$, we sample $r\%$ of tokens from DCLM-Edu and $(100-r)\%$ from bAbI. For bAbI, we ensure that each task has the same number of samples in the final dataset. As we continue pretraining from the base model without instruction tuning, we randomly interleave samples from both sources and segment them into sequences of 2048 tokens to match the model's context length.

We performed all experiments on a cluster with 8 nodes of $4\times$ A100 40GB GPUs. The largest experiments are around 32 GPU hours.
Here are the full hyperparameters we used in our main experiments:

\begin{table}[h]
    \centering
    \caption{Hyperparameters used in our main experiments.}
    \label{tbl:hyperparameters}
    \begin{tabular}{l|>{\centering\arraybackslash}p{3.cm}}
        \toprule
        \textbf{Parameter} & \textbf{SmolLM2-1.7B} \\
        \midrule
        Gradient Clip Val & 1.0 \\
        Precision & bf16-mixed \\
        Optimizer & AdamW \\
        Beta1 & 0.9 \\
        Beta2 & 0.95 \\
        Eps & $1.0 \times 10^{-8}$ \\
        Weight decay & 0.01 \\
        Learning rate & 0.0005 \\
        Lr Num warm-up Steps & 10\% \\
        Lr Decay Factor & 0.1 \\
        Lr Schedule & Cosine \\
        Flash Attention & True \\
        Context size & 2048 \\
        Batch size & 256 \\
        \bottomrule
    \end{tabular}
\end{table}

\newpage
\subsection{Hyperparameter Search}
\begin{figure*}[!htbp]%
    \centering
    \includegraphics[width=1.0\textwidth]{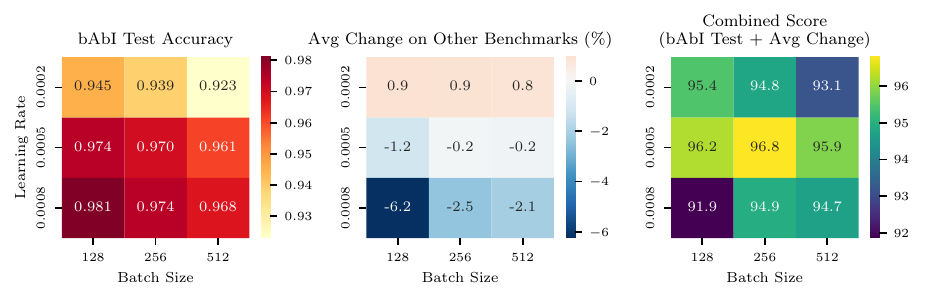}
    \caption{$3 \times 3$ grid search over learning rate and batch size. We used 100k bAbI-Synthetic samples with $50\%$ replay.}
    \label{fig:hp_grid_search}
\end{figure*}

We conducted a grid search to identify suitable hyperparameters for our main experiments. We searched over batch sizes \{128, 256, 512\} and learning rates \{0.0002, 0.0005, 0.0008\}. We used 100k samples from the bAbI-Synthetic dataset with a replay ratio of $50\%$ for the grid search, as this configuration demonstrated strong performance on the bAbI test set benchmark in preliminary experiments. We present the complete grid search results in Figure~\ref{fig:hp_grid_search}. We selected the configuration that achieved the highest combined score, which corresponded to a learning rate of 0.0005 and a batch size of 256.

\subsection{Dataset Statistics}
\begin{figure*}[!htbp]%
    \centering
    \includegraphics[width=1.0\textwidth]{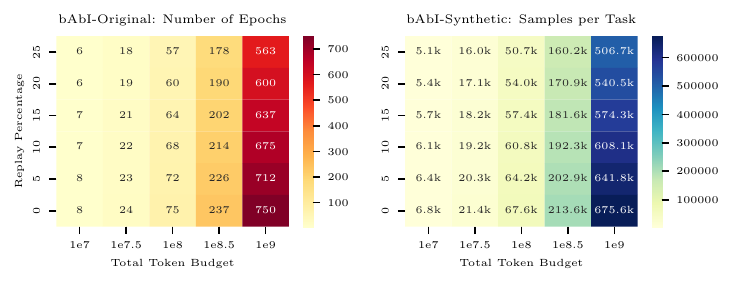}
    \caption{Number of epochs used for bAbI-Original and number of samples per task used for bAbI-Synthetic for each configuration.}
    \label{fig:epochs_vs_samples}
\end{figure*}

Figure~\ref{fig:epochs_vs_samples} shows the number of epochs used for bAbI-Original and the number of samples per task used for bAbI-Synthetic to reach the total token budgets for each replay percentage.

\newpage
\FloatBarrier
\section{Retaining GSM8K Performance}
\label{app:gsm8k}
\begin{figure*}[t]%
    \centering
    \includegraphics[width=0.85\textwidth]{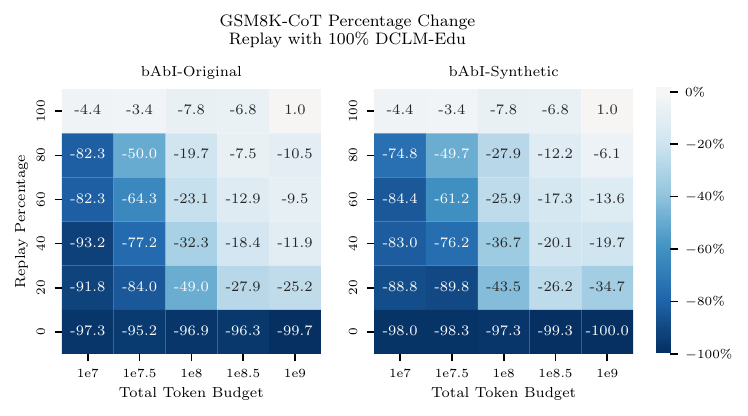}
    \caption{GSM8K-CoT (8-shot) performance utilizing the DCLM-Edu replay dataset across replay percentages of $20\%$, $40\%$, $60\%$, and $80\%$.}
    \label{fig:gsm8k_dclm_edu_only}
\end{figure*}

We initially also incorporated GSM8K-CoT (8-shot) within our suite of general performance benchmarks. Our analysis revealed that retention of GSM8K performance necessitates replay percentages exceeding $80\%$ when utilizing our selected replay dataset, DCLM-Edu. Figure~\ref{fig:gsm8k_dclm_edu_only} shows the results with replay percentages $20\%$, $40\%$, $60\%$, and $80\%$.

To investigate this, we conducted an analysis of GSM8K-CoT predictions from one of our trained models (1e8.5 total token budget, $40\%$ replay). We examined whether the trained model exhibited formatting issues that could lead to incorrect parsing of generated answers. However, we could not identify such parsing errors. Instead, the models trained on bAbI and DCLM-Edu appear to have experienced a genuine decline in mathematical reasoning capability. The following example illustrates this degradation:

\taskexample{
Q: James is putting together 4 tables. Each table has 4 legs and each leg needs 2 screws. He has 40 screws. How many screws will he have left over? \\
Target: 8 \\ \\
Base model response: "James has 40 screws. 4 tables need 4 legs and each leg needs 2 screws. 4 x 4 = 16 legs. 16 x 2 = 32 screws. 40 - 32 = 8. The answer is 8." \\ \\
Our trained model response: "James is putting together 4 tables. Each table has 4 legs. Each leg needs 2 screws.  He has 40 screws. So he will need 4 x 4 = 16 screws. He has 40 - 16 = 24 screws. The answer is 24."
}

We hypothesize that this degradation stems from the insufficient mathematical content within DCLM-Edu. \citet{allal2025smollm2} specifically perform upsampling of mathematical data in their original pretraining dataset mix. Consequently, DCLM-Edu alone proves insufficient to counteract the distributional shift induced by training on bAbI, except when employing very high replay percentages.

\begin{figure*}[t]%
    \centering
    \includegraphics[width=0.85\textwidth]{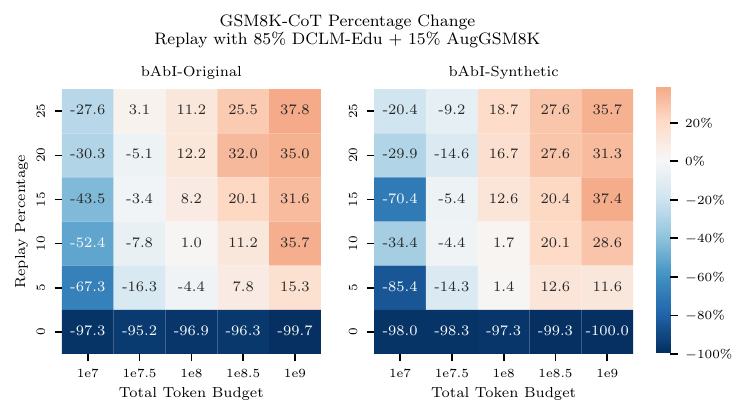}
    \caption{GSM8K-CoT (8-shot) performance utilizing a replay dataset comprising $85\%$ DCLM-Edu and $15\%$ AugGSM8K across total replay percentages of $5\%$, $10\%$, $15\%$, $20\%$, and $25\%$.}
    \label{fig:gsm8k_dclm_edu_and_auggsm8k}
\end{figure*}

To validate this hypothesis, we conducted additional experiments incorporating AugGSM8K~\citep{li2024mugglemath} samples within the replay mixture. Specifically, we allocated $15\%$ of the replay budget to AugGSM8K samples. The results are presented in Figure~\ref{fig:gsm8k_dclm_edu_and_auggsm8k}. The inclusion of AugGSM8K in the replay mixture not only completely mitigates the GSM8K-CoT benchmark degradation but also yields substantial performance improvements. Given that this performance enhancement would distort our combined score metric, we excluded this modified replay dataset configuration from our main experiments. Nevertheless, these findings demonstrate that for domains exhibiting significant performance degradation despite high replay percentages with a general replay dataset such as DCLM-Edu, allocating a modest proportion of the replay budget to domain-specific data effectively addresses the performance degradation.

\newpage
\FloatBarrier
\section{Full Benchmark Results}
\label{app:benchmarks}

\begin{figure*}[t]%
    \centering
    \includegraphics[width=1.0\textwidth]{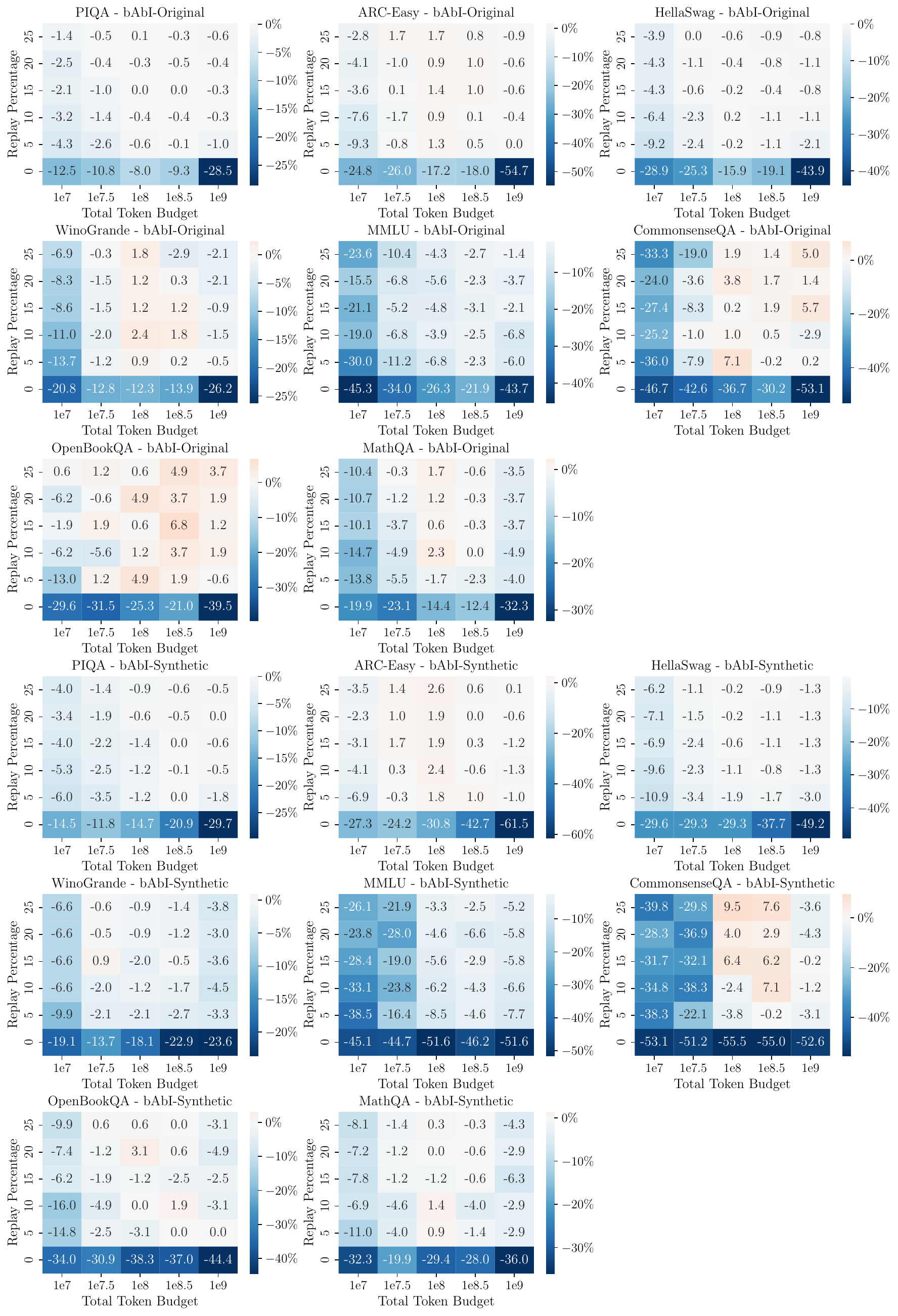}
    \caption{Full benchmark results of the individual general performance benchmarks of our main experiments. The first three rows show the results for bAbI-Original, the last three rows show the results for bAbI-Synthetic.}
    \label{fig:all_benchmarks}
\end{figure*}

Figure~\ref{fig:all_benchmarks} shows the full results of our main experiments for each individual benchmark for bAbI-Original and bAbI-Synthetic.

%%%%%%%%%%%%%%%%%%%%%%%%%%%%%%%%%%%%%%%%%%%%%%%%%%%%%%%%%%%%

\end{document}